\documentclass{article}
\usepackage{spconf,amsmath,graphicx}
\usepackage{booktabs}
\usepackage{tabularx}
\usepackage{multirow}
\usepackage{enumitem}
\usepackage[multiple]{footmisc} 
\usepackage{xcolor}

\usepackage{url}

\title{A Comparative Analysis of Machine Learning Approaches for Automated Face Mask Detection During COVID-19} 
\name{\normalfont{Junaed Younus Khan}\textsuperscript{*} \thanks{*The authors contribute equally to this paper.} \& Md Abdullah Al Alamin\textsuperscript{*}}
\address{junaedyounus.khan@ucalgary.ca, mdabdullahal.alamin@ucalgary.ca \\ \textit{University of Calgary}}

\begin{document}

\maketitle

\textbf{\textit{Abstract--} The World Health Organization (WHO) has recommended wearing face masks as one of the most effective measures to prevent COVID-19 transmission. In many countries, it is now mandatory to wear face masks, specially in public places. Since manual monitoring of face masks is often infeasible in the middle of the crowd, automatic detection can be beneficial. To facilitate that, we explored a number of deep learning models (i.e., VGG1, VGG19, ResNet50) for face-mask detection and evaluated them on two benchmark datasets. We also evaluated transfer learning (i.e., VGG19, ResNet50 pre-trained on ImageNet) in this context. We find that while the performances of all the models are quite good, transfer learning models achieve the best performance. Transfer learning improves the performance by 0.10\%--0.40\% with 30\% less training time. Our experiment also shows these high-performing models are not quite robust for real-world cases where the test dataset comes from a different distribution. Without any fine-tuning, the performance of these models drops by 47\% in cross-domain settings.}



\section{Introduction} \label{sec:intro}
The novel coronavirus disease (COVID-19) has created a global health crisis taking 3.7M lives and affecting $\sim$172M people. This is a highly contagious disease (more infectious than influenza and Ebola \cite{centers2020similarities, francis2020just}) and mostly transmitted through respiratory droplets. Hence, several precautionary measures are recommended to prevent its transmission e.g., social distancing, wearing mask, cleaning hands. Although vaccines are available, the worldwide mass vaccination is not yet done. It is also possible to be affected even after getting vaccinated since vaccines are not 100\% effective. While vaccinated people are less likely to have severe complications after getting COVID-19, a few exceptions are not unheard of\footnote{\url{https://globalnews.ca/news/8279892/covid-deaths-fully-vaccinated/}}. Moreover, due to genetic mutation, the virus is continuously evolving. The most recent strain ``Omicron" has already become a great concern for global health. Hence, precautionary measures are still crucial. In fact, wearing mask is considered the most effective measure according to WHO. Several studies ask for mass mask wearing since they found it to be highly beneficial to prevent COVID-19 \cite{cowling2009facemasks, eikenberry2020mask}.

It is found that crowded places spread the virus most  \cite{who_covid}. In many countries, government has made it mandatory to wear face mask in public places and public transportation. However, many people still show reluctance towards wearing masks due to negligence and lack of awareness \footnote{\url{https://www.bbc.com/news/health-58979626}}. Hence, it is necessary to locate unmasked people, make them aware and ensure mask wearing. Since, it is practically infeasible do it manually in the middle of a crowd or a public place, automated detection of face mask is probably the only way for this, and therefore currently receiving great attention from the research community \cite{inamdar2020real, bhambani2020real, chowdary2020face, loey2021hybrid,nagrath2021ssdmnv2}.


Most existing works in this direction suffer from any of these two limitations- \textbf{(1) Lack of Models Explored.} The number of evaluated models or techniques is limited, \textbf{(2) Lack of Datasets Used.} They did not evaluate the models on different dataset. To fill this gap, we explored a variety of models from three different perspectives (i.e., Baseline 1VGG, well-known VGG19 \& ResNet50, and transfer learning). We also evaluated the concept of domain adaptation in this context. In particular, we addressed two research questions.

\noindent
\textbf{RQ1. How accurately different models can detect face mask?}
We compared the performance of different models i.e., 1VGG, VGG19, ResNet50, and transfer learning (VGG19 \& ResNet50 pretrained on ImageNet) in mask detection task using two different datasets (Simulated and Real). It will help us to have a complete picture of the possibilities and challenges of automated face mask detection. In our study, the best performing model is ResNet50 with transfer learning which achieves 100\% \& 99.45\% accuracy in simulated \& real dataset respectively.  

\noindent
\textbf{RQ2. How effective the concept of domain adaptation is in face mask detection?} We evaluated domain adaptation by using simulated and real images as source and target domain respectively to check model transferability from simulated to real in this context. One motivation of doing so is to explore the performance of the models that are trained on simulated data but later deployed/tested on a different (real) dataset. We observed that currently there is a shortage of real images for incorrectly worn mask in the existing benchmark datasets (most of them are are simulated). Hence, this RQ will help us to evaluate the feasibility of detecting incorrectly masked images in real world using simulated data. Our experiment shows that model performance severely drops (by ~47\%) in such cross domain setting. However, the performance quickly improves when the trained model is fine-tuned using real data. This signifies that models that are trained on simulated dataset may not perform well on target real world dataset without some fine-tuning.

We believe that our comparative analysis will help the research community to explore further in this context and to select the most appropriate mask detection method.





\noindent

\textbf{Replication Package.} {\small Our code is shared at \url{ https://github.com/JunaedYounusKhan51/FaceMaskDetection}}



\section{Dataset} \label{sec:dataset}
We studied two datasets that are used for various other studies \cite{chowdary2020face, loey2021hybrid}.

\subsection{Simulated Facemask dataset (SFMD)} 
\begin{figure}[t]
\centering
\includegraphics[scale=0.2]{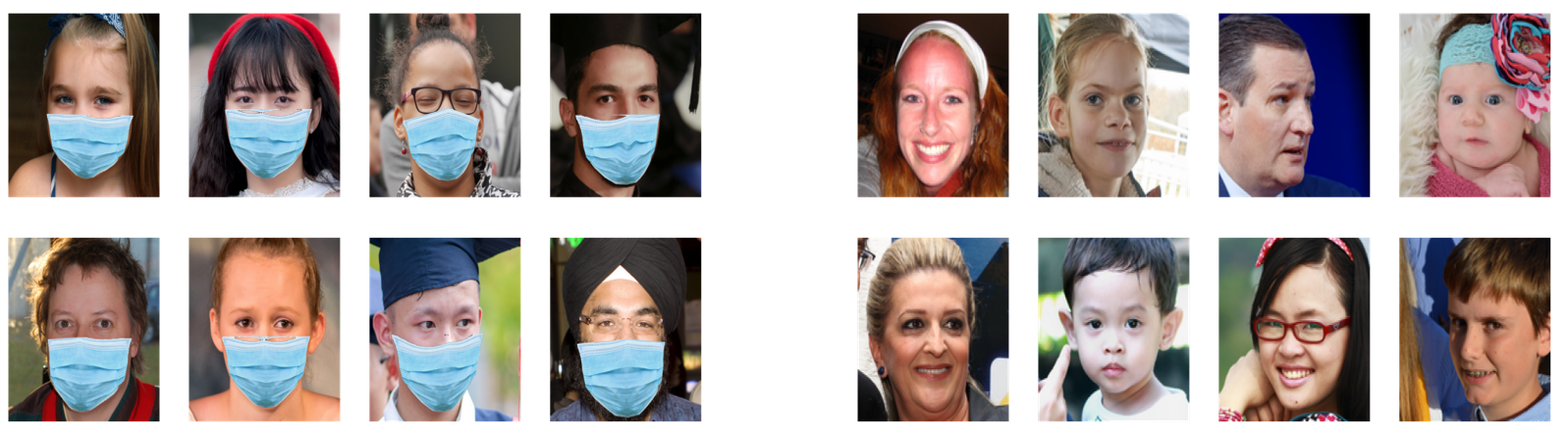}
\caption{Simulated Masked Face Dataset (SMFD) Samples}
\label{fig:simulated_dataset_samples}
\end{figure}

The dataset is published by Cabani et al. \cite{masknet}. It contains high resolution (1024×1024) 67K simulated samples of people wearing mask correctly and incorrectly. The original images are collected from Flicker dataset \cite{kazemi2014one}. For our study, we randomly selected 6,442 masked images from SFMD and  6,442 unmasked images from the Flicker dataset (Table \ref{fig:dataset_summary}). Some examples are available in Figure \ref{fig:simulated_dataset_samples}.

\subsection{Real Facemask dataset (RFMD)}
\begin{figure}[t]
\centering
\includegraphics[scale=0.24]{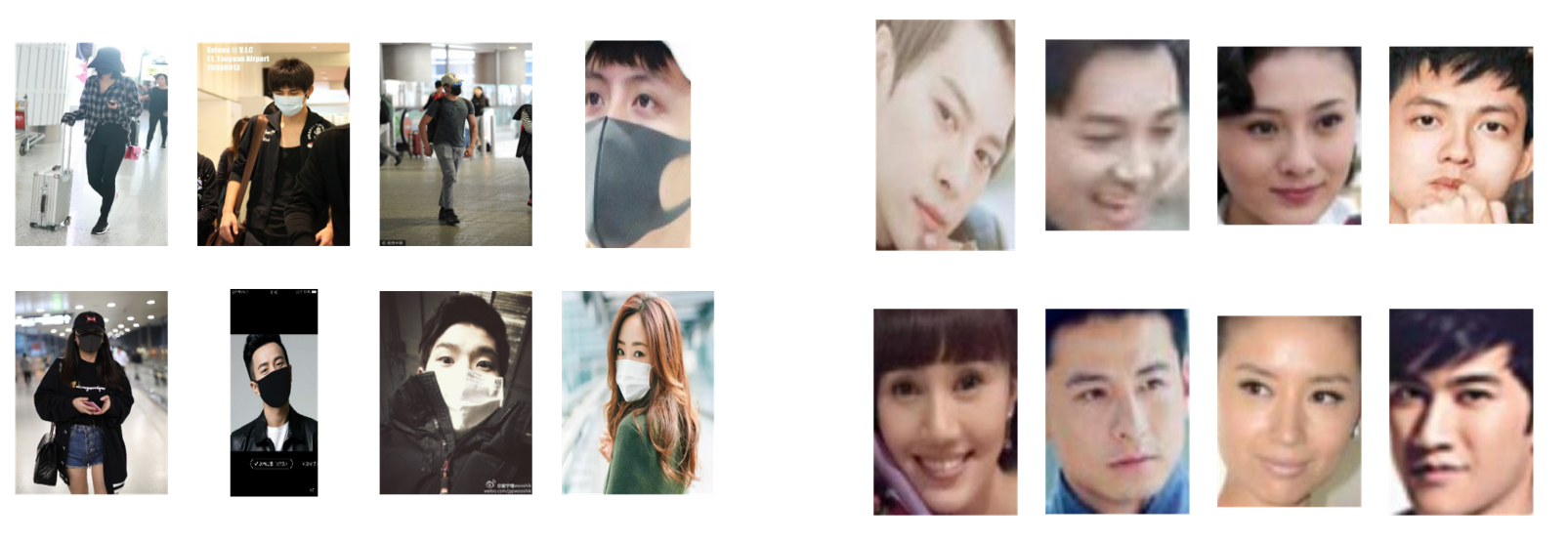}
\caption{Real Masked Face Dataset (RMFD) Samples}
\label{fig:real_dataset_samples}
\end{figure}

This is the biggest dataset of masked face for this problem \cite{real_facemask}. It consists of 90K images of 525 different people without masks and 6,442 images of real masked people. Among the masked images, many are from the same 525 people and some are different. Figure \ref{fig:real_dataset_samples} shows some samples from this dataset. To keep a balanced dataset, we used 6,442 masked and 6,442 unmasked images as described in Table \ref{fig:dataset_summary}.

\subsection{Train, Test and Validation split} 
 As mentioned, each of our datasets now contain 12,884 images (6,442 masked and 6,442 unmasked). We split the dataset into train and test set maintaining 90:10 ratio. We further divided the train portion to prepare train and validation set in 88:12 ratio. Hence, for both dataset, 12,884 images were divided into 10,203 train, 1,291 validation, and 1,288 test images.

\begin{table}[!htp]\centering
\caption{Overview of the facemask datasets}\label{tab: }
\resizebox{\columnwidth}{!}
{
\begin{tabular}{l|rr|rr}\toprule
\textbf{} &\multicolumn{2}{c}{\textbf{Original }} &\multicolumn{2}{c}{\textbf{Our Study}} \\\cmidrule{2-5}
\textbf{Dataset} &\textbf{With Mask} &\textbf{Without Mask} &\textbf{With Mask} &\textbf{Without Mask} \\\midrule
Real Facemask & 6442 & 90000 & 6442 & 6442 \\
\midrule
Simulated Facemask & 67049 & 65534 & 6442 & 6442 \\
\bottomrule
\end{tabular}
}
\label{fig:dataset_summary}
\end{table}

\section{Experimental Setup} \label{sec:methodology}
\begin{figure}[t]
\centering
\includegraphics[scale=0.33]{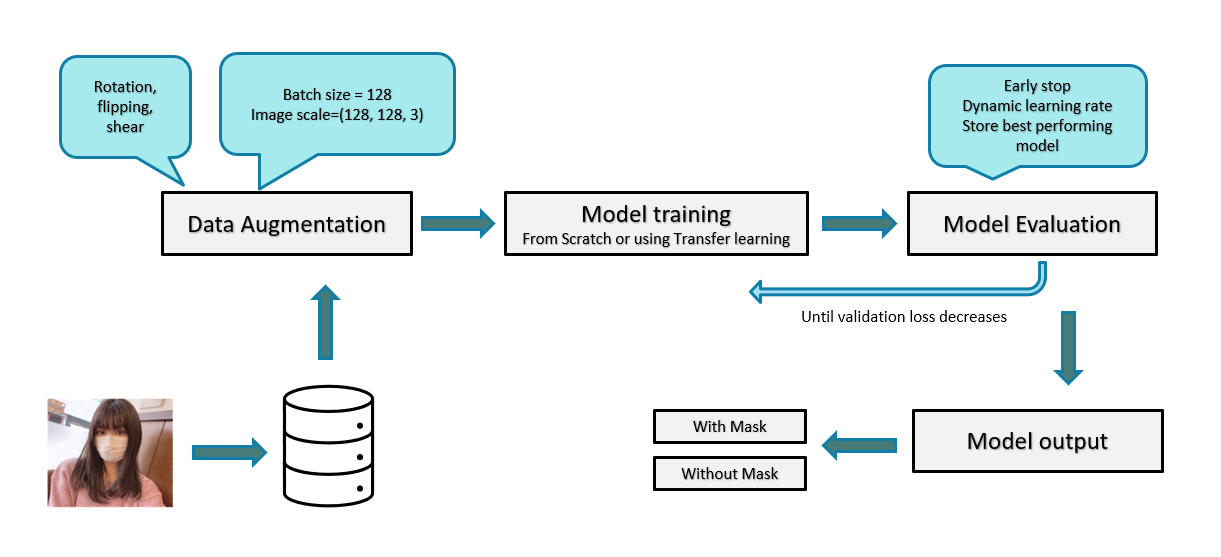}
\caption{Overview of our research methodology}
\label{fig:research_overview}
\end{figure}

In this section, we describe our experimental setup. An overview of our methodology is showed in Figure \ref{fig:research_overview}.

\subsection{Data Augmentation}
\begin{figure}[t]
\centering
\includegraphics[scale=0.25]{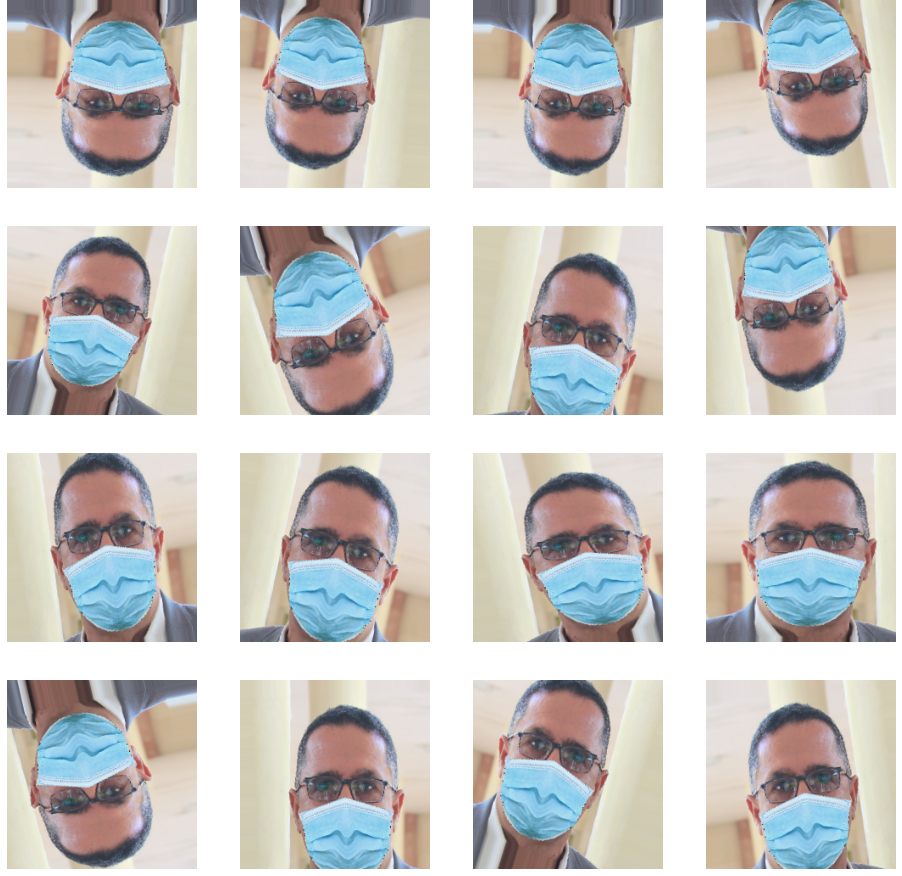}
\caption{Samples from augmented dataset}
\label{fig:data_augmentation_samples}
\end{figure}

We used different data augmentation techniques such as rotation (20\textdegree), smearing (0.2), flipping (horizontal \& vertical), zooming (range 0.1) as showed in Figure \ref{fig:data_augmentation_samples}.
Our dataset images were of different sizes (e.g., 1024×1024, 128×128). For data augmentation we used batch size of 128 and converted image dimension to 128×128.

\subsection{Models}

\textbf{\indent Baseline.} As a baseline model, we used a basic VGG block. First, we used a normalized layer of the size of the input images (e.g., 128X128X3), then we used 32 filters of size 3×3 followed by a max pooling layer. For the simplicity of this model, we did not use any Dropout for regularization.

\textbf{VGG19.} It is a CNN that is 19 layers deep. We trained it from scratch. We used Keras's API to get the VGG architecture and changed the final output layer from 1000 to 2.

\textbf{ResNet50.} We also experimented with ResNet50 architecture, a CNN containing 50 layers. Similar to VGG19, we trained ResNeT50 model from scratch.

\textbf{Transfer Learning.} To evaluate transfer learning, we used pre-trained VGG19 and ResNet50 (on ImageNet). We discarded the top layer and froze the base model. Then we flattened the output and used a dense layer with Softmax activation.

\subsection{Domain Adaptation} \label{subsec:domain_adaptation}
Domain adaptation is the ability to apply a model trained in one  domain (source ) to a different but similar domain (target). We evaluated domain adaptation in the context of mask detection where we considered the simulated dataset as the source domain and the real dataset as the target domain. We trained our best model on the simulated data and test it on real data without retraining. Later, we fine-tuned the model on real dataset and observed the change in performance.

\subsection{Training Process}
We initialized each model with Adam Optimizer and initial learning rate of 0.0001. As this is a binary classification, we used categorical cross entropy as loss function and accuracy as a metric to be observed. We started the training with 300 epochs and defined four callbacks to prevent over-fitting. 1) We stored the best model after each epoch. 2) We stopped the training if the validation loss does not decrease for three epochs. 3) We half the learning rate after 10 epochs. 4) We stop the training if the validation accuracy was 100\%. All models are trained on TALC cluster of University of Calgary with NVIDIA Tesla T4 GPU.

\subsection{Evaluation Metrics}
We reported the performances using standard evaluation metrics. Accuracy ($A$) is the ratio of correctly predicted instances out of all the instances. Precision ($P$) is the ratio between the number of correctly predicted instances and all the predicted instances for a given class. Recall ($R$) is the ratio of the number of correctly predicted instances and all instances for that class.{\scriptsize
\begin{eqnarray*}
A = \frac{TP+TN}{TP+FP+TN+FN},~
P  = \frac{TP}{TP+FP},~
R = \frac{TP}{TP+FN}
\end{eqnarray*}}

\begin{table}[!htp]
\caption{Performance of the Deep learning models}\label{tab:result}
\resizebox{\columnwidth}{!}
{
\begin{tabular}{l|rrr|rrr}\toprule
\multirow{2}{*}{\textbf{Model}} &\multicolumn{3}{c}{\textbf{Simulated Dataset}} &\multicolumn{3}{c}{\textbf{Real Dataset}} \\\cmidrule{2-7}
&\textbf{A} &\textbf{P} &\textbf{R} &\textbf{A} &\textbf{P} &\textbf{R} \\\midrule
Baseline-1VGG &99.8 &99.8 &99.8 &98 &98 &98 \\
VGG19 &99.8 &99.8 &99.8 &99.2 &99.2 &99.2 \\
ResNet50 &99.6 &99.6 &99.6 &96.8 &96.8 &96.8 \\
TL-VGG19 &99.7 &99.7 &99.7 &98.6 &98.6 &98.6 \\
TL-ResNet50 &\textbf{100} &\textbf{100} &\textbf{100} &\textbf{99.45} &\textbf{99.45} &\textbf{99.45}\\

\bottomrule
\end{tabular}
}
\end{table}

\section{Result} \label{sec:result}
In this section, we answer two research questions:

\noindent
\textbf{RQ1.} How accurately different models can detect face mask?
\noindent
\textbf{RQ2.} How effective the concept of domain adaptation is in face mask detection?

\subsection{Performance of ML models (RQ1)}
Table-\ref{tab:result} shows the summary of the result of the experiment. From the table, we see that all the models perform moderately well in both datasets, however, the best performing model is ResNet50 with transfer learning. It achieves an accuracy of 100\% in the simulated dataset and 99.45\% in the real dataset.

\subsection{Domain Adaptation Result (RQ2)}

For this RQ, we did not consider any transfer learning model (that is already pre-trained on ImageNet) so that we can figure out the actual adaptation capability of model in the context of face mask detection. Hence, we trained VGG19 (our best performing non-transfer learning model) on the simulated dataset (source) and test it on the real dataset (target). It achieves 52\% accuracy (i.e., 47\% performance drop). Note that the performance of this same model is 99.2\% when it is trained and tested on the real dataset (Table \ref{tab:result}). Then we took the model (trained on source dataset) and fine-tuned it on using target dataset. We find that the performance has quickly increased ($\sim$99\%) after the fine tuning. 

\begin{figure}[t]
\centering
\includegraphics[scale=0.16]{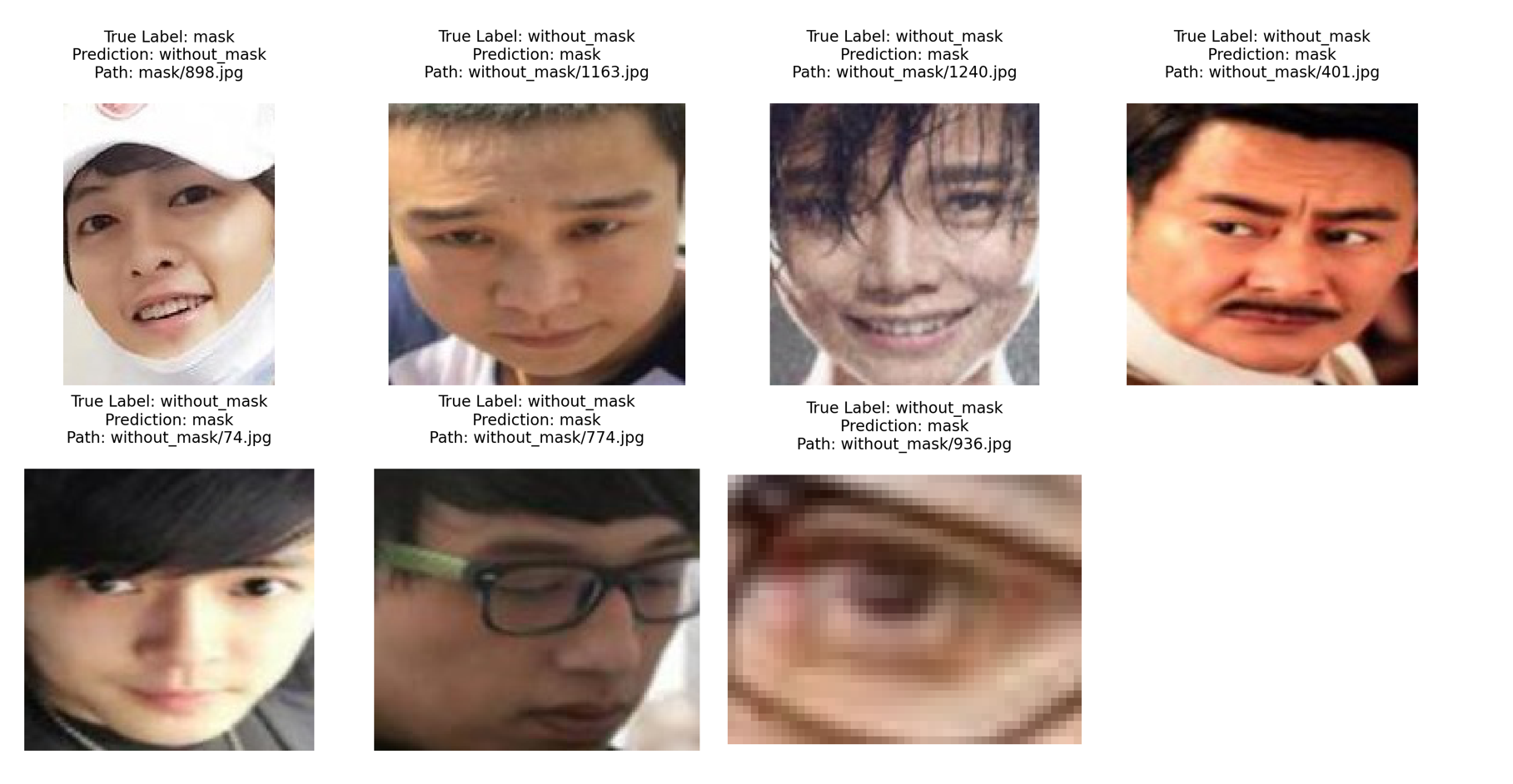}
\caption{TL-ResNet50 mispredicted samples on RMFD}
\label{fig:resnet50_misprediction}
\end{figure}

\begin{figure}[t]
\centering
\includegraphics[scale=0.19]{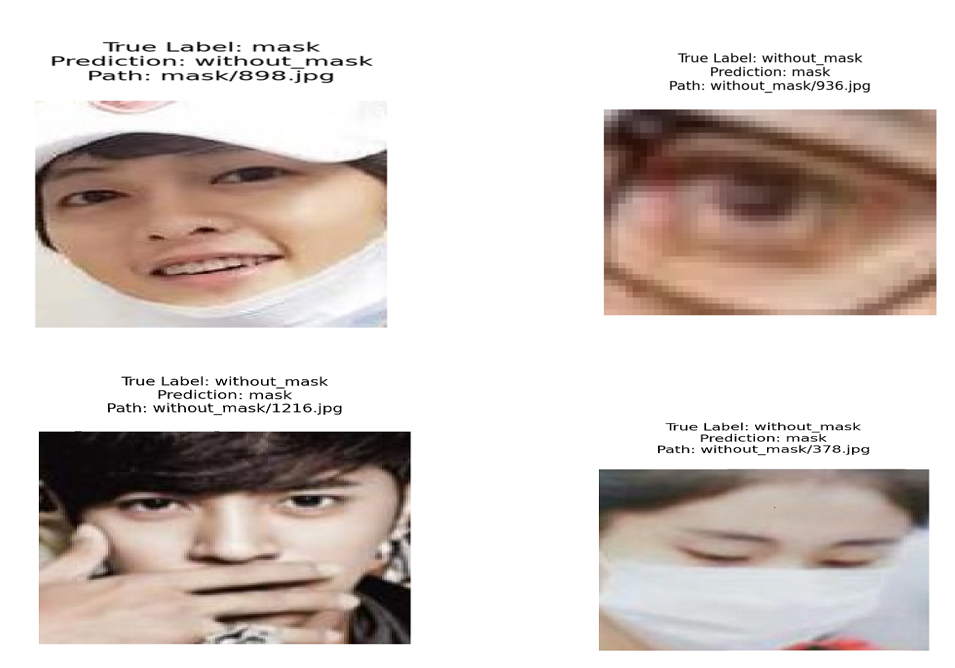}
\caption{Difficult/incorrectly labeled test samples on RMFD}
\label{fig:incorrect_label_real_image}
\end{figure}

\subsection{Error Analysis}
We find that most models perform very good in both datasets. Figure \ref{fig:resnet50_misprediction} shows the 7 incorrectly predicted samples by our model. From our misclassification analysis and data exploration, we can see there are some samples in real test data which are either incorrectly labeled or very difficult to predict (Figure \ref{fig:incorrect_label_real_image}). This shows, any model which is performing more than 99\% is one of the best possible models for this dataset.

\section{Discussion} \label{sec:discussion}

\subsection{Time Complexity Analysis of Different Models}
\begin{figure}[t]
\centering
\includegraphics[scale=0.68]{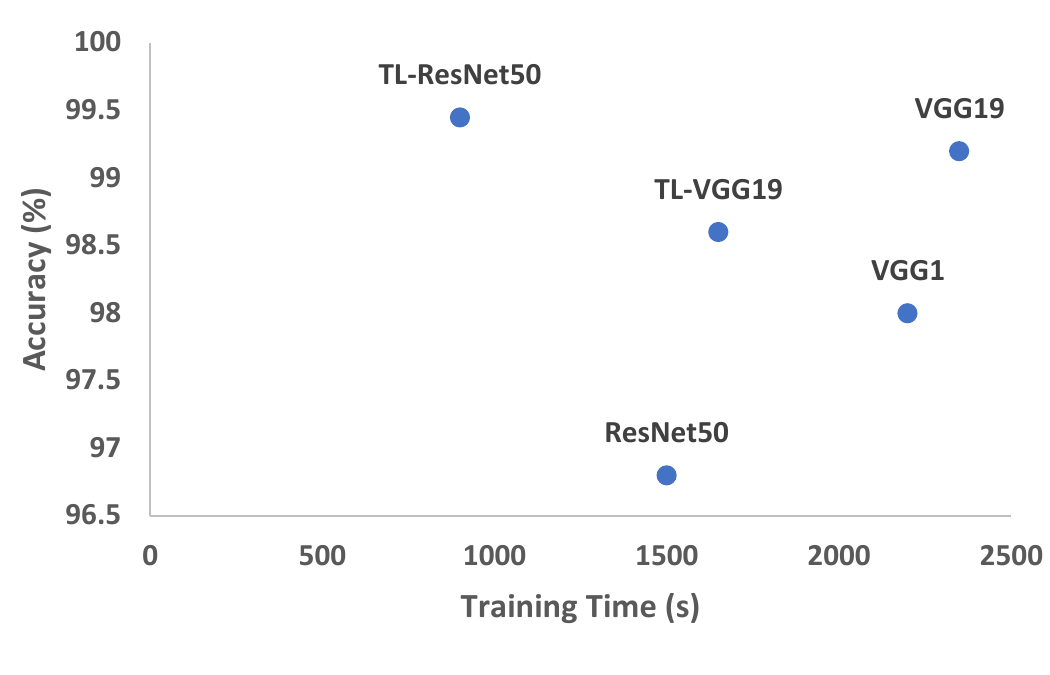}
\caption{Time and performance analysis of different models}
\label{fig:time_analysis}
\end{figure}

Besides performance, training time is an important concern for any ML model. Hence, we conducted a time complexity analysis of different models. Figure \ref{fig:time_analysis} shows the performance and required training time for every model on real dataset. We can see that transfer learning (e.g., TL-ResNet50) can achieve high performance (99.45\%) with very short training time ($<$1000s). Nearest competitor in terms of training time, ResNet50, achieves 96.8\% accuracy in 1500s.


\subsection{Implications of the Study}
All models performed very well, even the simple baseline achieved $>$98\% in both datasets. It has two implications-- (1) Binary classification of masked vs unmasked is probably a simple problem (at least for used datasets), (2) Even a simple model like VGG1 can be successfully used for some problems with proper data augmentation. Another lesson could be the effectiveness of transfer learning. We can see how transfer learning improved the performance with less training time. 

\subsection{Limitations}
The unmasked images of the real dataset mainly focused on the face only, but the masked images sometimes have whole body and multiple people as well. This phenomenon can have some manipulative effect on the model performance. Also, we did not make sure not to have same people in train and test set. However, as all the 525 persons of unmasked images are also present in masked images and the images are moderately diverse as well, there is little chance of bias here.


\section{Related Work} \label{sec:related_work}
ML algorithms can help prevent the spread of the COVID-19 pandemic in many ways such as forecasting \cite{rustam2020covid}, diagnosis \cite{minaee2020deep}, exploring potential drugs for re-purpose \cite{khan2020toward}. Recently, several studies have also focused on using ML techniques for automated face mask detection. Some of them demonstrated solution to detect face mask in real time camera images\cite{nagrath2021ssdmnv2, inamdar2020real, bhambani2020real}. Some proposed solution to detect if face mask is worn correctly or not ~\cite{tomas2021incorrect}. Chowdary et al. proposed a transfer learning approach by fine-tuning InceptionV3 for face mask detection ~\cite{chowdary2020face}. They used a different version of simulated dataset which contains around 700 masked and 700 without masks images. Loey et al. proposed a hybrid model where they used a pre-trained Resnet50's 2nd last layer as the feature vector for classical ML algorithms such as SVM, decision trees ~\cite{loey2021hybrid}. Both of these studies report more than 99\% accuracy.
\section{Conclusion} \label{sec:conclusion}

Our study shows that the current DL models for image classification are powerful for mask detection. Specially, performances of pre-trained transfer learning models are promising considering their time effectiveness. Future works can focus on detecting mask in real time and incorrectly worn mask. We also see that model's performance drops significantly on real data when it is trained on simulated data. This indicates that the models are not robust for such real life usage.


\bibliographystyle{abbrv}
\bibliography{Main}

\end{document}